\documentclass[11pt]{extarticle}
\usepackage[english]{babel}
\usepackage[utf8]{inputenc}
\usepackage[maxbibnames=99,backend=biber,style=authoryear]{biblatex}
\DeclareNameAlias{author}{family-given} 
\usepackage{csquotes}
\usepackage{fancyhdr}
\usepackage{graphicx}
\usepackage{amsmath}
\usepackage{amsthm}
\usepackage{amssymb}
\usepackage{amsfonts}
\usepackage{enumerate}
\usepackage{float}
\usepackage{authblk}
\usepackage{caption}
\usepackage{booktabs}
\usepackage{tabularx, ragged2e}
\usepackage{subcaption}
\usepackage{lipsum}
\usepackage{listings}
\usepackage{hhline}
\usepackage{hyperref}
\usepackage{hanging}

\newcolumntype{R}{>{\raggedright\centering\arraybackslash}X}
\newcolumntype{S}{>{\raggedright\centering\arraybackslash\hsize=.5\hsize}X}

\usepackage[dvipsnames]{xcolor}
\usepackage[dvipsnames]{xcolor}

\definecolor{Event}{HTML}{006400}
\definecolor{External}{HTML}{00008b} 
\definecolor{Solidarity}{HTML}{b03060}
\definecolor{Person}{HTML}{ff0000}
\definecolor{Policy}{HTML}{ecd440}
\definecolor{News}{HTML}{deb887}
\definecolor{Football}{HTML}{00ff00}
\definecolor{Spread}{HTML}{00ffff}
\definecolor{Italy}{HTML}{ff00ff} 
\definecolor{None}{HTML}{a6a6a6}
\newcolumntype{P}[1]{>{\centering\arraybackslash}p{#1}}
\title{Exploring Spatial-Temporal Variations of Public Discourse on Social Media: A Case Study on the First Wave of the Coronavirus Pandemic in Italy}
\author[1]{Michael Anslow}
\author[1]{Martina Galletti} 
\affil[1]{Sony Computer Science Laboratories, 75005 Paris, France}

\providecommand{\keywords}[1]
{
  \small	
  \textbf{\textit{Keywords---}} #1
}

\begin{document}
\captionsetup[table]{font=small,skip=4pt}
\captionsetup[figure]{font=small,skip=4pt}
\captionsetup[subfigure]{font=small,skip=4pt}

\maketitle
\begin{abstract}



This paper proposes a methodology for exploring how linguistic behaviour on social media can be used to explore societal reactions to important events such as those that transpired during the SARS CoV2 pandemic. In particular, where spatial and temporal aspects of events are important features. Our methodology consists of grounding spatial-temporal categories in tweet usage trends using time-series analysis and clustering. Salient terms in each category were then identified through qualitative comparative analysis based on scaled f-scores aggregated into hand-coded categories. To exemplify this approach, we conducted a case study on the first wave of the coronavirus in Italy. We used our proposed methodology to explore existing psychological observations which claimed that physical distance from events affects what is communicated about them. We confirmed these findings by showing that the epicentre of the disease and peripheral regions correspond to clear time-series clusters and that those living in the epicentre of the SARS CoV2 outbreak were more focused on solidarity and policy than those from more peripheral regions. Furthermore, we also found that temporal categories corresponded closely to policy changes during the handling of the pandemic. 

\end{abstract}


\section{Introduction}

SARS Cov2 spread rapidly across the world, resulting in immense suffering and death among those afflicted with severe cases of the disease. Subsequent emergency health measures to curtail the spread of the disease also changed the day-to-day lives of many people. This was true, even though the spread of SARS Cov2 itself was often unevenly geographically distributed, both nationally and internationally. These characteristics of the outbreak require that both spatial and temporal features are available in data concerning the events that transpired and explored in the analysis of this data to adequately understand it. To this end, we propose that textual social media data is a suitable resource for investigating this phenomenon, so long as location information and timestamps are available. Social media data is both voluminous and available across much of the globe on various social media platforms, allowing for timely updates and broad geographic coverage. To analyse this phenomenon, we propose a methodology that consists of identifying differences in term usage among tweets aggregated into different temporal and spatial categories already grounded in data. This is a general approach that could be applied to the study of various phenomena. To exemplify this, we apply our methodology to a compelling case study.

The fight against the SARS Cov2 epidemic relies partly on individuals' acceptance of self-distancing policies and understanding of the tragic nature of the pandemic. Differences in responses to the calamity and embracing of legislatures can be influenced by individuals' locations. Studies in psychology underlined how an individual's spatial distance can result in differences in the mental visualisation of an event~(\cite{CLT}). This was shown to be particularly true for the language encoding of an event~(\cite{Fujitapsychology}; \cite{Soderberg}; \cite{hayward}). In particular, experiments~(\cite{Fujitapsychology}) proved that people tend to use a more abstract and less concrete language when referring to an event that happened far away from them. To our knowledge, such hypotheses had never been explored in correlation with the outburst of a pandemic and in the context of social media. 

We used the analysis of Italian tweets in Italy as a case study to fill this gap. Italy was one of the first countries in Europe to be affected by SARS Cov2, the virus outbreak was unevenly distributed across the country with the main epicentre being in the north of Italy. Major towns in Lombardy, such as Bergamo, Cremona, Lodi, and Brescia registered an increase in excess mortality of 568\%, 391\%, 370\%, and 290\%~(\cite{Istat:2020}) respectively in the month of March 2020 and the north had an overall regional excess mortality of 185\%~(\cite{Istat:2020}). On the opposite end of the severity spectrum, during the same period, the south and central regions experienced excess mortality of 9\% and 13\% respectively~(\cite{Istat:2020}). Initially, the Italian government attempted to contain the outbreak in the north of Italy, with local lockdown measures, however, a national lockdown was soon adopted making the effects of the SARS Cov2 outbreak felt nationwide. Differences in the degree of intensity with which the virus was present in regions affected how Italians felt and reacted to the SARS Cov2 pandemic. 

The structure of the remainder of this paper is as follows: we first detail relevant existing prior work. Key characteristics of the data being used is then described, including preprocessing steps taken before analysis. Following this, we detail how spatial-temporal categories are grounded in data and how distinctive but representative terms are identified for each category using scaled f-scores. These terms are then hand-coded into term categories that capture major term types and their proportions across spatial-temporal categories are studied. Finally, we show that our approach captures salient and sensible spatial-temporal information. Moreover, the notion of ``epicentre'' and ``periphery'' posited by psychological studies, are shown to be empirically grounded in our data, and key differences between the two spatial categories are identified. 


\section{Literature Review}

Many researchers have turned to Twitter social media analysis to investigate reactions to the SARS Cov2 pandemic~(\cite{Pennycook}; \cite{Ferrara_2020}; \cite{Samuel_2020}, \cite{li2020depressed}). This type of data is a voluminous and instantaneous account of user reactions, capturing moment to moment snapshots of what happened and what is happening across the globe. Such data also spanned back to the very start of the virus outbreak as with~\cite{Chen_2020}, allowing for analysis at every stage of the spread of the pandemic. Nevertheless, the feeling of the general public or specialised groups on social media was rarely studied in connection with spatial and temporal variations. The only study which we could find was the one of~\cite{Su2020} which explored the differences in lockdown impacts in Wuhan and Lombardy. Their conclusions underlined that if in Lombardy people focused on leisure activities and the stress generally decreased, in Wuhan, there was an increase in attention of religion and emotions. If this study was a step further in exploring the connection between emotional engagement and linguistic behaviour during a pandemic, it did not explore the value of spatial distance in exploring the reactions to a major event, focusing on the two major epicentre of the SARS Cov2 pandemic.  

In psychology, different studies had been dedicated to the effects of spatial distance on the encoding of a particular event. If already~\cite{hayward} studied the commonalities between linguistic and visual representations of space, ~\cite{Soderberg} and, generally, the literature related to Construal Level Theory (see ~\cite{CLT}) underlined the relationship between the level of abstraction and psychological distance. \cite{Fujitapsychology} took a step further by studying the relationship between spatial distance and language encoding. Their experiments demonstrated that individuals tend to use more abstract language to describe spatially distant events.



\section{Data Processing}

The data consisted of tweets retrieved from the Twitter API via their tweet identifiers. These tweet identifiers were the product of the continuous and ongoing collection of tweets concerning the SARS Cov2 outbreak since its discovery in China as detailed in~\cite{Chen_2020}. It is worth noting that the data collection strategy changed over time as the pandemic propagated, for example, initially ``the novel coronavirus'' was one of the initial search terms for collecting relevant tweets, and later the term ``COVID19'' was added as an additional search term as it became more popular. It should also be noted that the set of tweet ids provided in the data set were for tweets available at the time that they were initially crawled. As such, there will always be some variation in which tweets were available depending on when tweets were requested by their identifiers using the Twitter API. However, these tweets made up a negligible fraction of the data set.

Tweets contain diverse types of structures that are adopted for multitudinous reasons. Some of this is explicitly baked into Twitter, such as referring to particular users using ``@'' and the user name to ``mention'' another user. Other content is stylistic, like using many repeated hashtags or arranging hashtags in a particular way. Our focus was to try and capture the content of what was being communicated rather than who was doing the communicating or whom it was communicated to. As such we filtered out mentions, URLs, and boilerplate text. This boilerplate content was the text that was repeated across multiple tweets that had nothing to do with the content being communicated. For more information on how we processed the text see the Text Processing Subsection~\ref{section:textprocessing} in the Appendix. We should note that the same methodology but with changes to which terms are filtered could be used to also target different aspects of tweets' content, such as identifying shared websites or popular users.

\section{Spatial \& Temporal Categories}

\begin{figure}[!ht]
  \centering
     \begin{subfigure}[b]{0.48\textwidth}
         \centering
         \includegraphics[width=\textwidth]{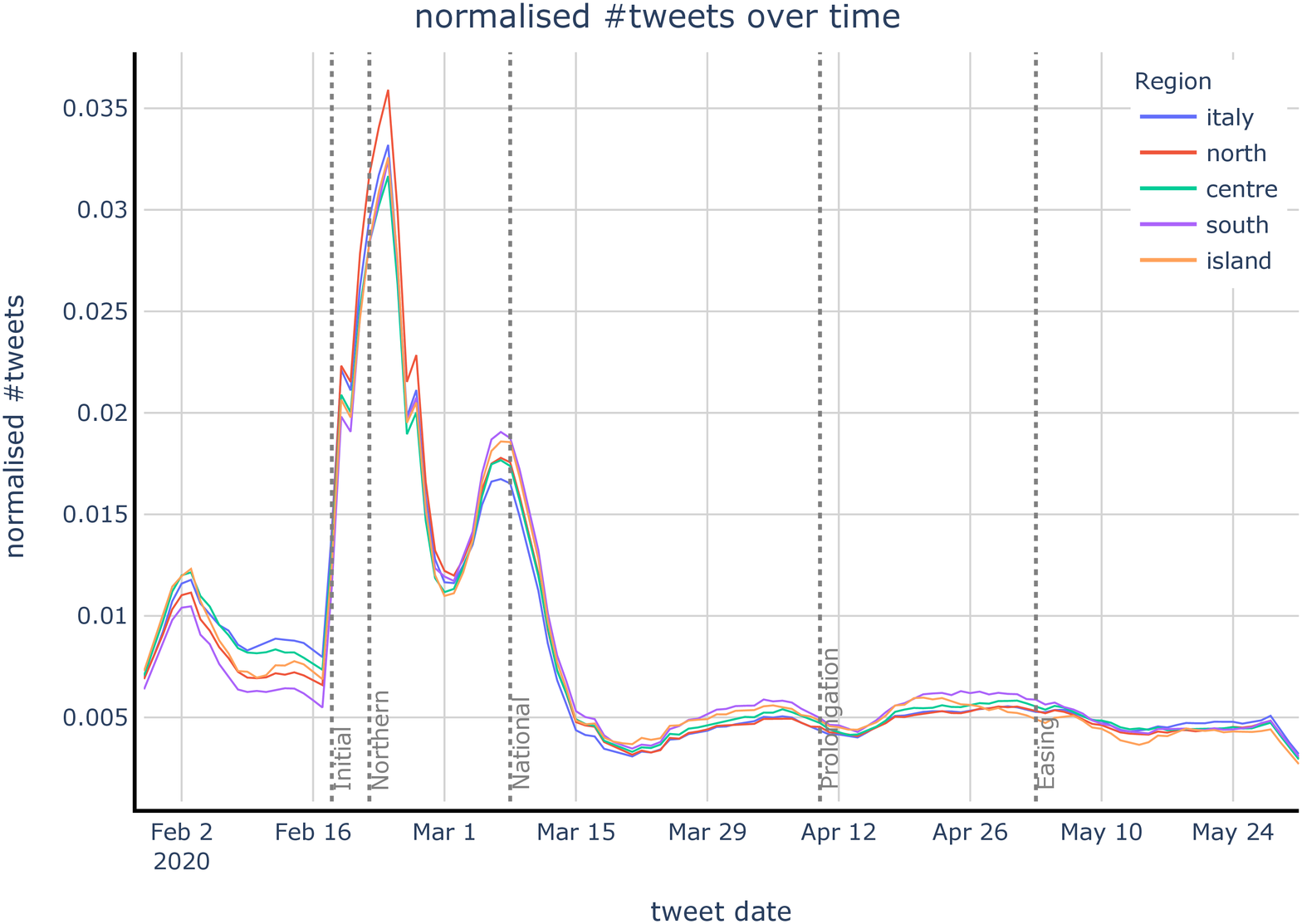}
         \caption{}
     \end{subfigure}
     \centering
     \begin{subfigure}[b]{0.48\textwidth}
         \centering
         \includegraphics[width=\textwidth]{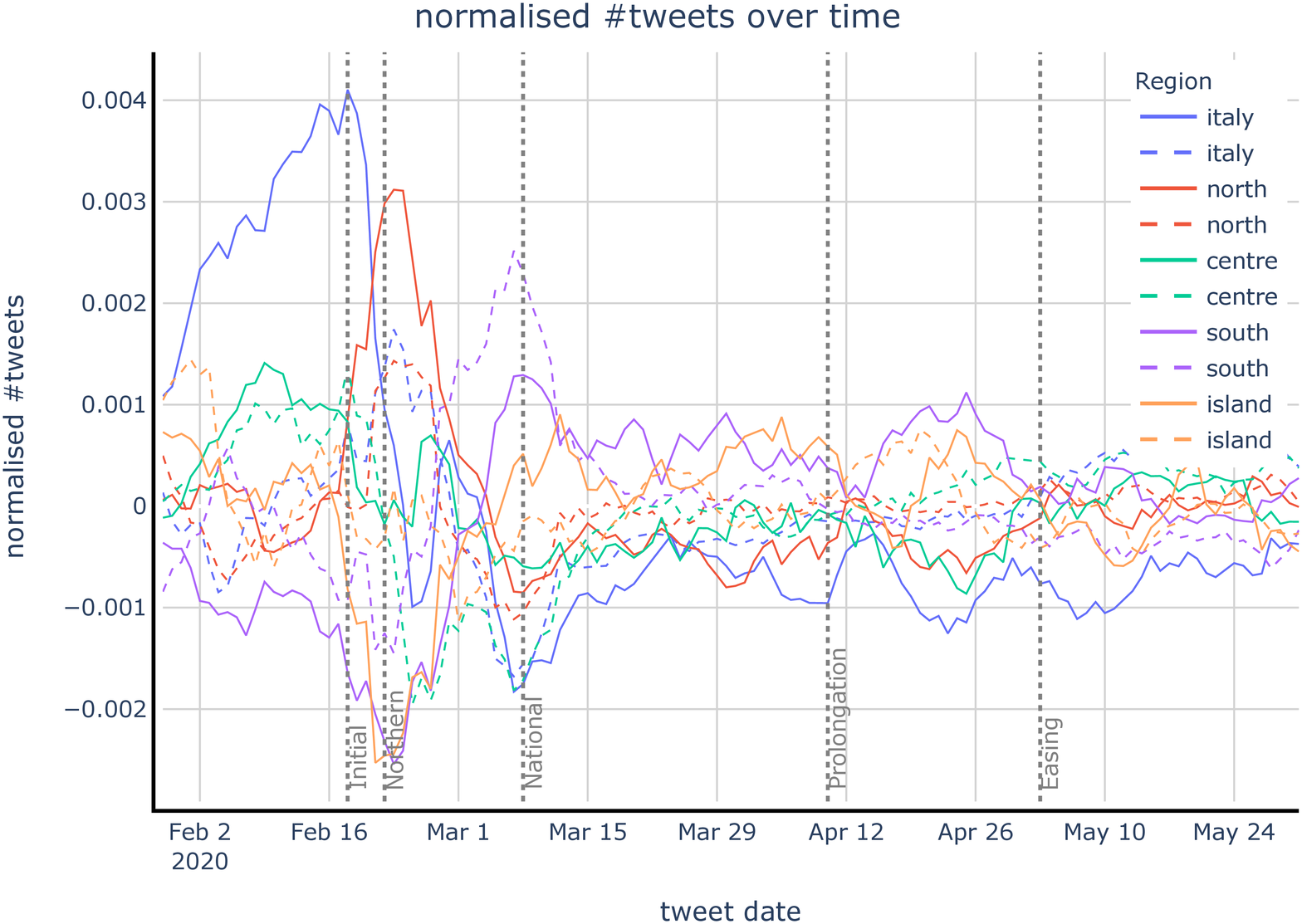}
         \caption{}
     \end{subfigure}
     \hfill
     \begin{subfigure}[b]{0.48\textwidth}
         \centering
         \includegraphics[width=\textwidth]{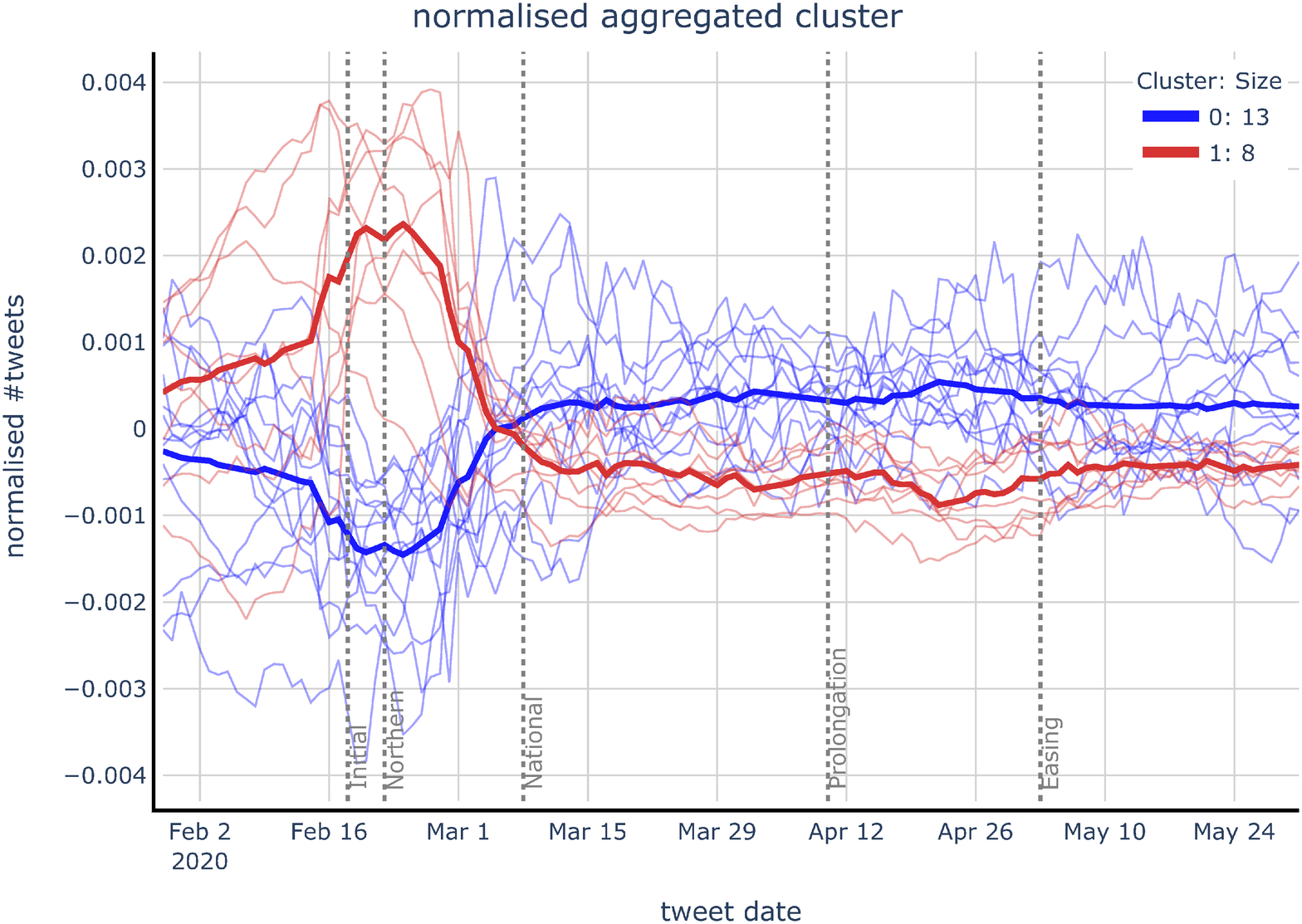}
         \caption{}
     \end{subfigure}
     \hfill
    \caption{(a) Normalised \#tweets for coarse categories. (b) Trend removed normalised  \#tweets for original (solid) and retweeted (dashed) tweets. (c) Aggregated time-series clusters. 0: Calabria, Sardinia, Basilicata, Sicily, Campania, Puglia, Marche, Friuli-Venezia Giulia, Umbria, South, Trentino-Alto Adige, Abruzzo, Molise 1: Lazio, Lombardy, Italy, Piedmont, Emilia Romagna, Liguria, Tuscany, Veneto.
    }
        \label{fig:coarse_trends}
\end{figure}

There were 1.3 million tweets in the data set which we could map onto both coarse-grained and fine-grained Italian regions. The mapping procedure is described in the Normalising Locations Subsection~\ref{section: locations} in the Appendix. The regions had the following share of tweets: ``North'' (which included both ``North East'' and ``North West'' regions) 36\%,``Italy'' 24\%, ``Centre'' 24\%, ``South'' 10\%, ``Islands'' 6\%. The generic ``Italy'' region consisted of tweets where finer-grained categories such as ``South'' or ``Centre'' could not be determined. The normalised number of tweets tweeted across regions is shown in Figure ~\ref{fig:coarse_trends}.a. We can see that all regions had very similar tweet trends. This indicated a similar intensity of the reaction to the SARS Cov2 outbreak at similar events in the trends. However, this did not indicate the content of ``what'' was tweeted, only that these regions responded to particular events with similar, but not identical, intensity.

The events in Figure~\ref{fig:coarse_trends}.a provided a natural partition of the data into time periods. By cross-referencing these terms with historical records, we found that events in the time series corresponded very closely to key moments of policy change. From this information we constructed five different periods for subsequent investigation as follows: 
\begin{enumerate}[\indent {}]
   \item {\bf Pre}. By ``Pre'' we designated the period starting from the end of January until the 18th of February. During this period, no Italian cases had been detected yet. The only cases were, in fact, imported ones (two Chinese tourists in Spallanzani hospital). 
   \item {\bf Initial}. ``Initial'' indicated a very short, yet significant period, included between the 19th of February and the 22nd of February when the first Italian cases were detected in the village of Codogno and Veneto region.
   \item {\bf Northern}. With ``Northern'' we referred to the period from the 23rd of February to the 9th of March when the first strict measures were being applied to only some areas in the ``North'', firstly the ``red zones'' and later, entire regions.
   \item {\bf National}. ``National'' indicated the month when a national lockdown was imposed, starting from the 10th of March and ending on the 10th of April.
   \item {\bf Prolongation}. By ``Prolongation'' we meant the two weeks starting from the 10th of April, when the lockdown was extended, until the 5th of May. Moreover, the modalities of the relaxation of the measures were announced during those weeks.
   \item {\bf Relaxing}. With this we referred to the ease of the lockdown, starting from the 5th of May until the end of the month. 
\end{enumerate}

Though we did not automatically identify these periods, it was apparent from the time-series that the raw number of tweets was a good proxy for identifying reactions to policy change. The two major peaks corresponded to the regional lockdown and the national lockdown. We could also see smaller bulges approximately peaking at the lockdown prolongation and finally the lockdown easing. These could easily be segmented using automatic time-series segmentation algorithms based on the peaks in the data. The ``Initial'' period was arguably not needed, but we believed that it was an interesting period to capture in our analysis as it caught the shift from the SARS Cov2 being a largely Asian problem to it becoming an Italian and, later on, a European-wide problem. We chose not to create another time period based on the peak at around February the 2nd. We believe it might correspond to the outbreak on the Diamond Princess Cruise Ship~(\cite{pmid32207674}) and as our focus was largely on Italy we didn't feel it was necessary to investigate this in more detail. Instead, it was captured in the ``Pre'' period.

In Figure~\ref{fig:coarse_trends}.b, the normalised mean trends across all coarse regions were subtracted from the trend for each region. This was done per tweet type (original and retweet). This accentuates whether particular regions were tweeting more or less than the general trend of Italy per tweet type. 

We could see three waves of activity in the data. First ``Italy'', consisting of mostly original tweets and to a lesser extent, ``Centre'' in the ``Pre'' period. There was then a large peak in the ``North'' during the ``Initial'' period, which was unsurprising given that the latter occurred in those areas. The ``South'' then peaked during the ``Northern'' period, perhaps as concern grew that a national lockdown was eminent. Significantly, a large number of these tweets were retweets. A plausible explanation for this is that the ``South'' was sharing events happening in the ``North'', as that was where most of the news concerning the spread of the virus was originating from. Finally, the ``Islands'' largely followed the mean of ``Italy'', though with more activity in the ``Pre'' period.

It is a puzzle why Twitter users identified their location as ``Italy'' and ``Centre'' more in the ``Pre'' period. One might expect ``Italy'' to have the Italian average and ``Centre'' to be somewhere between the ``North'' and the ``South''. One explanation of why this was not the case might be to do with international identity. In the case of those identifying their location as ``Italy'' rather than their particular residing region, might indicate less-localised identity and more focus on global events. Similarly, the ``Centre'' also contained Rome, the capital of Italy, which might have a similar tendency to focus on international events.

In Figure~\ref{fig:coarse_trends}.c, all Italian regions were clustered using K-Means time-series clustering with euclidean distance and the number of clusters selected via Silhouette score. This gave a clear distinction between regions responding more during the ``Initial'' period and regions responding more during the ``National'' one. These clusters roughly reflect the regions consisting of the epicentre of the SARS Cov2 outbreak (``Cluster 1'') and other regions (``Cluster 0''). From now on, we will refer to these clusters as the ``Epicentre'' and the ``Periphery''. This provided a natural spatial separation that was empirically grounded in the data that will be used in further analysis. Interestingly, our a priori spatial categorisation into Italian regions (``North'', ``Centre'', ``South'', ``Islands'') collapsed to only two clusters with very distinctive trends. The fact that this naturally corresponded to the ``Epicentre-Periphery'' hypothesis we are exploring is itself some evidence that the phenomenon might be supported by the data.
 
\section{Epicentre vs. Periphery}

Our primary point of investigation was to identify differences between the ``Epicentre'' and the ``Periphery'' over different temporal periods. For this purpose, we used the scaled f-score (SFS) proposed by~\cite{kessler2017scattertext} that was developed specifically to identify the most distinctive terms between two categories of textual documents. This score was calculated for each term in a class of documents based on Precision (how indicative this term is of a class vs. other classes) and Recall (how common this word is within a class). The calculation of this is shown in the Scaled F-Score Subsection~\ref{section: scaledfscore}. For each category, all tweets were aggregated together and contrasted with all tweets not belonging to that category. The categories considered were: spatial-temporal, only spatial, and only temporal; the latter two were akin to marginalising out spatial or temporal categories. Further to this, we also hand-coded terms into broad term categories to allow for a more general analysis of the terms and to allow for quantification of term type ratios across categories. These term categories were as follow:

\begin{enumerate}
    \item \textcolor{Italy}{\textbf{Regions \& Locations inside Italy}}. This included any spatial granularity including buildings such as hospitals (for example the ``Lazzaro'' and ``Spallanzani'' hospital). It also included the people of an Italian place, such as ``Neapolitans''.
    \item \textcolor{External}{\textbf{Regions \& Locations outside Italy}}. This included any spatial granularity outside Italy, including Cruise Ships such as the ``Diamond Princess cruise ship''. It also incorporated people of a foreign place, such as the ``Chinese''.
    \item \textcolor{Event}{ \textbf{Events/Times}}. This included particular events such as ``Easter'', or broad periods of time, such as ``July'' or ``Autumn''. No events related to particular policies, such as the date of the national lockdown were in this term category. 
    \item \textcolor{Solidarity}{\textbf{Solidarity}}. Terms related solidarity with policy measures, such as ``\#istayathome'' or more generally, solidarity with the situation people were in.
    \item \textcolor{Spread}{\textbf{Disease Spread \& Statistics}}. Terms related to the spread of the disease, such as the number of new cases and deaths, terms related to the severity of the disease, particular specific reports of new cases (for example, that of an infected   ``38-year-old'').
    \item \textcolor{Policy}{\textbf{Policy}}. Terms related to enacting policies, important policy dates, and the content of policies such as washing hands with alcoholic gel.
    \item  \textcolor{Person}{\textbf{People}}. Specific notable individuals related to SARS Cov2 outbreak.
    \item \textcolor{News}{\textbf{News}}. Indicators of news such as terms related to breaking news, and news updates.
    \item \textcolor{Football}{\textbf{Football}}. References to football such as football matches or football teams. This category superseded the generic locations if the main reason for the location being mentioned was related to a football match.
    \item \textcolor{None}{\textbf{Uncoded}}. Terms not belonging to any category.
\end{enumerate}

There was a methodological difficulty in justifying the category of a term given that much of the context for that term was lost. For example, seeing the term {\em bresciani}, `people of Brescia', might be an indication that the virus was spreading across the city of Brescia, which was one of the main epicentre during the first wave. However, looking more closely at the data, we understood that this was a reference to the football match between Brescia and Naples where the Brescia's fans shouted ``Naples Coronavirus'' as an insult. To overcome this, we directly searched for term context in the corpus and cross-referenced it with historical data before assigning a category to a term. As such, our hand-coding was very much based on expert knowledge of historic events and corpus study.

The hand-coded top 10 scaled f-score terms are shown in Table~\ref{table:fscorecomparison}. The terms have been translated into English while the original terms are shown in the Appendix, in Table 2~\ref{section: italiantable}. In our analysis, we focused on contextualising the most dominant term categories, rather than explaining the context of each term when these terms were not as significant. A complete glossary of all terms can be found in the Appendix, in the Glossary Subsection~\ref{section: glossarytable} that provides context for all terms in the table based on our exploration of the corpus and historical research. 

Firstly, looking across temporal categories, we can see that there was a common progression across clusters: 

\begin{table}[htp]
\tiny
\begin{center}
\setlength\tabcolsep{1.5pt} 
\begin{tabularx}{\textwidth}{SRRRS}
\toprule
\multicolumn{5}{c}{\textbf{The English Translation of the tokens obtained}} \\
\midrule
{} & Periphery & Epicentre & Marginal & Ratios\\
 \hline   
Pre          &   \textcolor{Spread}{that isolated}, \textcolor{Event}{\#sanremo2020}, \textcolor{News}{update}, \textcolor{Spread}{\#coronaviruschina}, \textcolor{Solidarity}{flash}, \textcolor{Spread}{suspected case}, \textcolor{Italy}{at cotugno hospital}, \textcolor{Italy}{battipaglia}, \textcolor{Spread}{first italian}, \textcolor{None}{\#sardine} &                                           \textcolor{Person}{niccolò}, \textcolor{Event}{\#sanremo2020}, \textcolor{External}{from wuhan}, \textcolor{External}{hong kong}, \textcolor{Italy}{\#spallanzani hospital}, \textcolor{Italy}{spallanzani hospital}, \textcolor{External}{hubei}, \textcolor{External}{china}, \textcolor{External}{\#chinese}, \textcolor{Policy}{isolato} &                                                                                                             \textcolor{Event}{\#sanremo2020}, \textcolor{Person}{niccolò}, \textcolor{External}{from wuhan}, \textcolor{Italy}{\#spallanzani hospital}, \textcolor{External}{hubei}, \textcolor{External}{\#chinese}, \textcolor{External}{egypt}, \textcolor{External}{hong kong}, \textcolor{Italy}{civitavecchia}, \textcolor{Italy}{spallanzani hospital} &  External:0.25, Spread:0.20, Italy:0.20, Event:0.10, News:0.05, Solidarity:0.05, Person:0.05, Policy:0.05 \\
Initial      &           \textcolor{Football}{people from brescia}, \textcolor{Football}{\#brescianapoli}, \textcolor{Football}{neapolitan}, \textcolor{Football}{choir}, \textcolor{Football}{they sing}, \textcolor{Italy}{cava}, \textcolor{Football}{neapolitans}, \textcolor{Italy}{\#codogno}, \textcolor{Spread}{contact with}, \textcolor{Italy}{codogno} &                                               \textcolor{Italy}{codogno}, \textcolor{Italy}{castiglione}, \textcolor{Spread}{38-year-old}, \textcolor{Italy}{\#lodi}, \textcolor{External}{princess}, \textcolor{Italy}{\#codogno}, \textcolor{Italy}{adda}, \textcolor{Spread}{38-year-old recovered}, \textcolor{Spread}{cases in lombardy}, \textcolor{Policy}{obligatory lockdown} &                                                                                                   \textcolor{Spread}{38-year-old}, \textcolor{Italy}{\#codogno}, \textcolor{Italy}{\#lodi}, \textcolor{Italy}{adda}, \textcolor{Spread}{38-year-old recovered}, \textcolor{Policy}{obligatory lockdown}, \textcolor{Spread}{cases in lombardy}, \textcolor{Italy}{in the lodi area}, \textcolor{Italy}{castiglione}, \textcolor{Spread}{infected in lombardy} &                                        Italy:0.40, Football:0.30, Spread:0.20, External:0.05, Policy:0.05 \\
Northern     &                                    \textcolor{Spread}{amuchina}, \textcolor{Policy}{closed schools}, \textcolor{Italy}{catania}, \textcolor{Person}{ceriscioli}, \textcolor{Policy}{summit}, \textcolor{Spread}{\#amuchina}, \textcolor{Italy}{\#palermo}, \textcolor{Italy}{palermo}, \textcolor{Italy}{from north}, \textcolor{Person}{emiliano} &                                                    \textcolor{Italy}{\#coronaviruslombardia}, \textcolor{Policy}{closed schools}, \textcolor{Policy}{doors}, \textcolor{Policy}{closed}, \textcolor{Spread}{amuchina}, \textcolor{Spread}{panic}, \textcolor{Policy}{\#redzones}, \textcolor{Football}{inter}, \textcolor{Spread}{\#coronavirus19italy}, \textcolor{Policy}{postponed} &                                                                                                                           \textcolor{Italy}{\#coronaviruslombardia}, \textcolor{Policy}{closed schools}, \textcolor{Spread}{amuchina}, \textcolor{Policy}{doors}, \textcolor{Policy}{closed}, \textcolor{Policy}{postponed}, \textcolor{Policy}{\#redzones}, \textcolor{Spread}{panic}, \textcolor{Policy}{schools}, \textcolor{Spread}{\#coronavirus19italy} &                                          Policy:0.35, Spread:0.25, Italy:0.25, Person:0.10, Football:0.05 \\
National     &  \textcolor{Solidarity}{\#westayathome}, \textcolor{Policy}{police}, \textcolor{Italy}{messina}, \textcolor{Italy}{syracuse}, \textcolor{Solidarity}{\#istayathome}, \textcolor{Person}{musumeci}, \textcolor{Italy}{\#sardinia}, \textcolor{Policy}{\#redzoneitaly}, \textcolor{Solidarity}{\#everythingwillbefine}, \textcolor{Italy}{benevento} &  \textcolor{Solidarity}{\#westayathome}, \textcolor{Solidarity}{\#istayathome}, \textcolor{Solidarity}{\#istayathome}, \textcolor{Policy}{self-certification}, \textcolor{Solidarity}{\#everythingwillbefine}, \textcolor{Policy}{\#redzoneitaly}, \textcolor{Event}{easter}, \textcolor{Solidarity}{\#stayathome}, \textcolor{Event}{april}, \textcolor{Spread}{\#coronavirus19italy} &                                                                              \textcolor{Solidarity}{\#westayathome}, \textcolor{Solidarity}{\#istayathome}, \textcolor{Solidarity}{\#istayathome}, \textcolor{Policy}{\#redzoneitaly}, \textcolor{Solidarity}{\#everythingwillbefine}, \textcolor{Policy}{self-certification}, \textcolor{Event}{easter}, \textcolor{Solidarity}{donate}, \textcolor{Solidarity}{\#stayathome}, \textcolor{Policy}{groceries} &                            Solidarity:0.40, Policy:0.20, Italy:0.20, Event:0.10, Person:0.05, Spread:0.05 \\
Prolongation &                                 \textcolor{Italy}{syracuse}, \textcolor{Policy}{\#4thmay}, \textcolor{Policy}{\#relatives}, \textcolor{Italy}{\#taranto}, \textcolor{Event}{april 25th}, \textcolor{Event}{\#25thapril}, \textcolor{Italy}{sassari}, \textcolor{Event}{\#easter}, \textcolor{Policy}{take away}, \textcolor{Policy}{for the phase} &                                                                                   \textcolor{Event}{easter}, \textcolor{Event}{may}, \textcolor{Policy}{\#mes}, \textcolor{Solidarity}{\#stayhome}, \textcolor{Spread}{nursing homes}, \textcolor{Policy}{reopen}, \textcolor{Event}{\#25thapril}, \textcolor{Policy}{\#phase2}, \textcolor{Person}{arcuri}, \textcolor{Policy}{phase} &                                                                                                                                                \textcolor{Event}{easter}, \textcolor{Solidarity}{\#stayhome}, \textcolor{Event}{\#25thapril}, \textcolor{Policy}{\#4thmay}, \textcolor{Event}{may}, \textcolor{Event}{\#easter}, \textcolor{Person}{arcuri}, \textcolor{Event}{april 25th}, \textcolor{Solidarity}{\#stayathome}, \textcolor{Policy}{phase 2} &                            Policy:0.40, Event:0.30, Italy:0.15, Solidarity:0.05, Spread:0.05, Person:0.05 \\
Relaxing     &                                         \textcolor{Spread}{covid free}, \textcolor{Italy}{fvg}, \textcolor{Italy}{\#bari}, \textcolor{Italy}{\#basilicata}, \textcolor{Italy}{\#puglia}, \textcolor{Italy}{\#sicily}, \textcolor{Italy}{caserta}, \textcolor{Spread}{zero infections}, \textcolor{Spread}{no death}, \textcolor{Italy}{basilicata} &                                                     \textcolor{Spread}{the bulletin of}, \textcolor{Person}{crisanti}, \textcolor{None}{webinar}, \textcolor{Policy}{during lockdown}, \textcolor{Person}{zangrillo}, \textcolor{Spread}{second wave}, \textcolor{External}{\#brazil}, \textcolor{Policy}{\#phase3}, \textcolor{Policy}{after the covid}, \textcolor{External}{brazil} &                                                                                                                        \textcolor{Italy}{mondragone}, \textcolor{Spread}{covid free}, \textcolor{Person}{zangrillo}, \textcolor{Event}{post covid}, \textcolor{Policy}{after the covid}, \textcolor{Policy}{after lockdown}, \textcolor{Spread}{zero deaths}, \textcolor{Policy}{\#phase3}, \textcolor{Spread}{no death}, \textcolor{Policy}{during lockdown} &                                          Italy:0.35, Spread:0.25, Policy:0.15, Person:0.10, External:0.10 \\
Clusters     &                                              \textcolor{Italy}{\#bari}, \textcolor{Italy}{\#taranto}, \textcolor{Italy}{\#abruzzo}, \textcolor{Italy}{\#sardinia}, \textcolor{Italy}{caserta}, \textcolor{Italy}{irpinia}, \textcolor{Italy}{\#basilicata}, \textcolor{Italy}{\#sicily}, \textcolor{Solidarity}{flash}, \textcolor{Italy}{salerno} &                                                                                           \textcolor{News}{\#latest}, \textcolor{Football}{\#asroma}, \textcolor{Italy}{liguria}, \textcolor{None}{\#della}, \textcolor{Italy}{\#lazio}, \textcolor{Italy}{romagna}, \textcolor{Italy}{florence}, \textcolor{Italy}{\#toscana}, \textcolor{Italy}{emilia}, \textcolor{Italy}{piemonte} &  \textcolor{Italy}{italy}, \textcolor{Spread}{emergency}, \textcolor{Spread}{dead}, \textcolor{Italy}{\#coronavirusitalia}, \textcolor{Policy}{lockdown}, \textcolor{Spread}{virus}, \textcolor{Spread}{cases}, \textcolor{Italy}{lombardy}, \textcolor{Spread}{contagions}, \textcolor{External}{china}, \textcolor{Italy}{\#covid19italia}, \textcolor{Spread}{positives}, \textcolor{Italy}{rome}, \textcolor{News}{update}, \textcolor{External}{chinese} &                                                                                                           \\
Ratios       &                                                                                                                                                                                                                                           Italy:0.39, Spread:0.17, Policy:0.14, Football:0.10, Event:0.07, Solidarity:0.07, Person:0.05, News:0.02 &                                                                                                                                                                                                                                                                           Policy:0.27, Spread:0.17, External:0.14, Italy:0.14, Event:0.10, Solidarity:0.10, Person:0.07, Football:0.02 &                                                                                                                                                                                                                                                                                                                                                                                                                                                               &                                                                                                           \\

\bottomrule
\end{tabularx}
\end{center}
\caption{Top 10 scaled f-scores words for the ``Epicentre'' and ``Periphery'' across varying periods. The ``Marginal'' row aggregates documents across time-periods categories, while the ``Marginal'' column aggregates documents across the ``Periphery'' and ``Epicentre'' categories. The bottom right corner, where the marginal row and column meet, is simply the most common terms across all documents. Finally, ratios of term categories are shown across time-periods and the ``Epicentre''-``Periphery''. These did not include the marginal terms. Uncommon terms found in this table are defined in the Glossary Subsection~\ref{section: glossarytable}}. 
\label{table:fscorecomparison}
\end{table}

\begin{enumerate}[\indent {}]
\item \textbf{Pre}. This period focused on the outbreaks outside of Italy in ``China'', ``Egypt'', and on the ``Diamond Princess Cruise ship''. There was also some mention of the first suspected Italian case from China in the ``Spallanzani'' hospital. A suspected case on the ``Costa Crociere'' cruise ship in the port of ``Civitavecchia'', near Rome, and one in the ``Cotugno'' hospital were mentioned. The ``Sanremo'' music festival was also a talking point.
\item \textbf{Initial}. This period focused on places within Italy and Football. The former could be found in further references to the hospital in ``Codogno'' where patient one, ``a 38-year-old'' was found and a focus on the ``Lombardy'' region, with references to specific towns such as ``Castiglione'' and ``Lodi.'' The latter referenced a football match between ``Brescia'' and ``Naples'' in which Brescia's fans chanted ``Naples Coronavirus''. We could also see initial talks in the ``Epicentre'' about an obligatory lockdown.
\item \textbf{Northern}. This period was dominated by policy-related terms and the spread of the SARS Cov2. We could see references to ``Amuchina'', the best known Italian hand sanitiser. The further policy acts included ``schools' closures'', first in the so-called ``red zones'' and later on in entire northern regions. We could also see that ``Lombardy'' continued to be a focus of discussion.
\item \textbf{National Lockdown} This period was dominated by messages of solidarity such as ``\#istayathome'', ``everythingwillbefine'' and ``wewillmakeit''. The policy continued to be an important category, with the first mention of a ``self-certification'' policy, with individuals having to fill out forms to leave their homes. 
\item \textbf{Prolongation} This period was mostly concerned with policy and events. The lockdown measures were extended and there was a lot of discussion about the end of the lockdown. After that date, people would be allowed to meet ``relatives''. ``Easter'', and Liberation Day on the ``25th of April'' were discussed as they could not be celebrated as usual. 
\item \textbf{Relaxing} There was a lot of discussion about the post-lockdown situation with terms referring to the ``after the lockdown'' and ``post lockdown'' period which officially would have started in ``June''. 
\end{enumerate}

There were important differences between the ``Epicentre'' and the ``Periphery''. While both spatial categories were equally concerned with the spread of the disease, the ``Epicentre'' focused more on policy while the ``Periphery'' focused on places in Italy where the outbreak was occurring. Also, the ``Periphery'' was completely unconcerned with regions external to Italy, while the ``Epicentre'' discussed it a lot. This might be because the disease was spread from the ``Epicentre'' to other regions, while this was not a concern in the ``Periphery''. It might also be that the ``Epicentre'' was focused on the more internationally connected north and centre of Italy, where Italians might have a different perspective of their relationship to the rest of the world. We could see that messages of solidarity were felt across both the ``Epicentre'' and ``Periphery'' during the National Lockdown. 

\section{Conclusions}

The temporal events in the time-series analysis corresponded well to key policy implementation events to curtail the SARS Cov2 outbreak. These events provided natural temporal categories for the exploration of Twitter data concerning the SARS Cov2 crisis. Given that we only performed a single case study, verifying this relationship between tweet intensity and policy implementation across different countries would be valuable for further exploration into reactions to SARS Cov2 policies in other countries.

The prominence of terms related to public health policy indicated that Twitter data may be a good resource for exploring public policy and discourse in future epidemic and pandemic crises. Furthermore, social media is perhaps the most effective way to measure this phenomenon as it is widely used and its content responds quickly to current events. As such, having a methodology to perform such an analysis along with historic case studies would be a valuable asset for planning public policy and discourse in future crises. Obviously, an important asset to analyse the enormous problem of policy acceptance in an emergency could be the development of a tool to automatically measure user engagement in pseudo-real-time towards particular policies by different nationalities and on several social media networks.

The non-uniform spread of SARS CoV2 in Italy had a different impact on people’s engagement and linguistic behaviour on Twitter. In fact, though there were very similar trends in tweets intensity across regions of Italy, we empirically grounded two broad spatial categories that were supported by the data. These spatial categories roughly corresponded to the ``Epicentre'' of the SARS Cov2 outbreak in the northern regions and other regions that made up the ``Periphery'' of the outbreak. This provided credence in the idea that there were indeed two different reactions between the ``Epicentre'' and the ``Periphery''.

Our temporal and spatial analysis on these categories showed that the main focus of tweets was on particular locations in Italy and outside of Italy where the disease was spreading, policy implementation, symbols of solidarity, the spread of the disease, particular people, and, to a lesser extent, news and football. These followed a trend of initially focusing on affected locations outside of Italy, particular Italian individuals and regions of Italy, policy, solidarity, key dates and events, and finally a diminishing spread of the disease.  

Differences in the psychological spatial distance corresponded to different linguistic behaviours towards the pandemics. In fact, there were distinctive differences between tweets from the ``Epicentre'' and the ``Periphery''. The majority of the focus in the ``Epicentre'' seemed to have been on policy measures and messages of solidarity, while the other regions focused on affected regions and particular people. However, we also saw in the trends of original tweets and retweets that the ``South'' of Italy, which was a large part of the ``Periphery'', was retweeting significantly during the national lockdown. As only original tweets were considered for text analysis, it might be the case that the South was retweeting stories originating from the ``Epicentre''. Exploring the content and origin of retweets would be helpful in better understanding the differences between the ``Epicentre'' and the ``Periphery''. 



We do not pretend to have conducted an exhaustive study or to have identified a definitive methodology for data analysis of this type. In particular, we restricted our analysis to one country and one social media network. Moreover, our analysis is agnostic to which linguistic phenomena it captures, in the sense that it is not targeting a particular feature of language use. However, it is easy to apply our methodology and use it to gain a helpful broad snapshot of a phenomenon in question to support further analysis.

\section{Bibliography}
\printbibliography

\section{Appendices}

\subsection{Text Processing}
\label{section:textprocessing}

Tweet text was tokenised using the Italian Spacy ``it\_core\_news\_sm'' model (\cite{spacy2}). This was chunked into bi-grams and tri-grams. We did not use lemmatisation as conjugation is an important feature of Romance languages and we found that a lot of context was lost by doing so. 
For example, {\em conti} is the plural for `counts' in Italian, while {\em conte} is not only the singular for `count', but also the surname of Prime Minister Giuseppe Conte. Obviously, we did not want to lemmatise {\em conti} to {\em conte} and to generate confusion with the surname of the Italian Prime Minister. 
Bi-grams and Trigrams were selected via thresholds set on both their Point-wise Mutual Information (PMI) and their frequency on texts aggregated by month. This aggregation allows for terms particular to a period of time to have more chances of being included as a n-gram. Thresholds for PMI and frequency were both set according to ranked cumulative density cut-off thresholds. So both PMI scores and frequency counts were ranked by score and count respectively from high-to-low and normalised to sum to 1. The cut-off thresholds were then selected by choosing the first value after the cumulative density reached predefined thresholds. This threshold was 0.75 for PMI (or 75\% of the PMI mass) and 0.15 (or 15\% of the frequency mass) for frequency. As PMI can contain negative values, the scores were shifted to start from 0 and the threshold chosen in the same way, only the original unshifted value was used for the term following the cumulative density threshold. 

We then performed further filtering and normalisation of the text as follows:

\begin{enumerate}
    \item Manually removing boiler-plate text. That is the text that was part of a fixed template used across tweets that did not contribute to Tweet meaning. For example, tweets that always started with the name of a news agency.
    \item Mapping uninformative n-gram variations, such as {\em in Italia} or {\em nell'Italia}.
    \item Mapping variations of ``Covid'' and ``Coronavirus'' to a single form such as ``Covid-19'' $\longrightarrow$ ``Covid19'' for both hashtag and word terms.
    \item Removing user mentions and URLs.
\end{enumerate}

\subsection{Normalising locations}
\label{section: locations}
To identify tweet locations, we relied on the self-reported user location associated with tweets. Clearly, this was not always truthful: if all reported locations were true, Antarctica would be a highly populated continent. However, we assumed the majority of users that chose to report their location as a ``real'' location, accurately did so. Based on this assumption we, first of all, normalised the users' self-identified locations using a hand-curated dictionary mapping. This mapping was created assigning all reported locations to standardised Italian regions if a location could be identified. For example, both {\em Milan, Lombardia} or {\em Lombardia, Milano} would both correspond to {\em Lombardia, Italia} in the normalised version of the data. 

Following the initial normalisation, we aggregated the data into coarser categories: for example, all northern locations (towns, villages, or cities) were aggregated into ``North, Italy''. These coarser categories allowed us to compare an established categorisation against the empirically grounded categories we identified later in the paper. We based our division between ``North'', ``South'', ``Centre'', and ``Islands'' on the NUTS~(\cite{EUreport:2007}) study of the European Union which grouped the twenty Italian regions into five geopolitical supra-regions ( ``North-East'', ``North-West'', ``Islands'', ``South'' and ``Centre'') of a purely statistical nature(\cite{EUreport:2007} p. 139). We decided to aggregate ``North-East'' and ``North-West'' supra-regions into a unique macro-category (``North'') because at the beginning the virus spread at the same time both in the ``North-West''( typically in Lombardy) ``North-East'' (for example in Veneto). 
On the contrary, the supra-region ``Islands'' seemed to be particularly important given the topographical nature of these territories, geographically separated from the rest of Italy and the spread of the virus.

\subsection{Scaled F-Score}
\label{section: scaledfscore}

The f-score takes the harmonic mean of these two scores with a $\beta$ parameter controlling the weighting towards precision ($\beta < 1$), towards recall ($\beta > 1$), or $\beta = 1$ a harmonic average of the two. As there is a general bias towards frequent words, even with $\beta < 1$, scores $z \in Z$ where $|Z|$ is the size of the vocabulary are transformed to their cumulative density value given a normal distribution with $\mu$ as the mean of the scores and $\sigma$ as the variance of the distribution as follows. 

\[
  \Phi(z) = \int_{-\inf}^{z} \mathcal{N}(\mu|z,\,\sigma^{2}) dx\,.
\] 

This reduces the impact of very frequent words the f-score.

\subsection{Glossary}
\label{section: glossarytable}

\begin{enumerate}[\indent {}]
    \itemsep0em 
    \item \textbf{25th of April}. Italian independence day.
    \item \textbf{4th May}. The first deadline set by the Italian government for easing the lockdown
    \item \textbf{Amuchina}. Amuchina indicates a series of products intended for environmental sanitation and disinfection, as well as industrial and body cleaning. The Amuchina hand gel is the best known.
    \item \textbf{Arcuri Domenico}. Domenico Arcuri is the extraordinary commissioner for the implementation and coordination of the measures necessary for the containment and contrast of the SARS CoV2 emergency. 
    \item \textbf{Ceriscioli Luca}. Luca Ceriscioli is the president of the Marche region.
    \item \textbf{Codogno}. A small village in Lombardy where the ``patient 1'' was detected in February.
    \item \textbf{Cotugno}. A very famous hospital in Naples.
    \item \textbf{Crisanti Andrea} Andrea Crisanti is an Italian professor of Microbiology at the University of Padua. He became famous during the pandemics for his analysis of citizens in Vò where he found that most of the infected people were asymptomatic carriers. 
    \item \textbf{Emiliano Michele}. Michele Emiliano is the president of the Puglia region and former mayor of Bari.
    \item \textbf{Inter}. Famous Italian team.
    \item \textbf{Meccanismo Europeo di Stabilità (MES)}. It is an international organization that originated as a European financial fund for the financial stability of the Euro-zone, which was to serve as a permanent source of financial assistance for member states in difficulty. The European Union has decided to open a 240 billion credit line for direct and indirect healthcare costs linked to the COVID-19 emergency for member states.
    \item \textbf{Mondragone} Mondragone is a small town near Caserta where an epidemic outbreak was identified in the Bulgarian agricultural workers' community leading to social and political tensions.
    \item \textbf{Musumeci Nello}. President of Sicilian region.
    \item \textbf{Niccolò} Niccolò Bizzarri, 21 years old, disabled, felt from his wheelchair because of a pothole in the street in Florence and he died in hospital on the 15th of January.
    \item \textbf{Phase 2}. Easing of the lockdown given a series of conditions which are met: transmission stability, non-overloaded health services, ability to promptly test all suspected cases, ability to ensure adequate resources for contact tracing, isolation, and quarantine. 
    \item \textbf{Phase 3}. Return to normality which only a vaccine and a cure will give.
    \item \textbf{Sanremo 2020/ Sanremo 70}. A Traditional Italian music event held at the beginning of February
    \item \textbf{Sardine}. Sardine is a movement that organises rallies all over Italy to protest against Salvini's policies. The demonstrations brought together a total of 300,000 people in Milan, Florence, Naples, and Palermo. This movement does not belong to any political party and is moreover mainly made up of young people.
    \item \textbf{Spallanzani} The Hospital in Rome where the first cases, two Chinese tourists, were detected.
    \item \textbf{Zangrillo Alberto}. Professor in Anaesthesiology and Intensive Care at San Raffaele university and Head of Anaesthesia and Intensive Care Unit of IRCCS Hospital San Raffaele in Milan. He is also the private doctor of Silvio Berlusconi. 
    \item \textbf{Zingaretti Nicola}. President of the Lazio region. 
\end{enumerate}

\subsection{Italian Results}
\label{section: italiantable}

\begin{table}[H]
\tiny
\begin{center}
\setlength\tabcolsep{1.5pt} 
\begin{tabularx}{\textwidth}{SRRRS}
\toprule
\multicolumn{5}{c}{\textbf{The Italian tokens obtained}} \\
\midrule
{} & Periphery & Epicentre & Marginal & Ratios\\
\hline
Pre          &             \textcolor{Spread}{che ha isolato}, \textcolor{Event}{\#sanremo2020}, \textcolor{News}{update}, \textcolor{Spread}{\#coronaviruschina}, \textcolor{Solidarity}{flash}, \textcolor{Spread}{caso sospetto}, \textcolor{Italy}{al cotugno}, \textcolor{Italy}{battipaglia}, \textcolor{Spread}{primo italiano}, \textcolor{None}{\#sardine} &                                                                \textcolor{Person}{niccolò}, \textcolor{Event}{\#sanremo2020}, \textcolor{External}{da wuhan}, \textcolor{External}{hong kong}, \textcolor{Italy}{\#spallanzani}, \textcolor{Italy}{spallanzani}, \textcolor{External}{hubei}, \textcolor{External}{china}, \textcolor{External}{\#cinesi}, \textcolor{Policy}{isolato} &                                                                                                                                    \textcolor{Event}{\#sanremo2020}, \textcolor{Person}{niccolò}, \textcolor{External}{da wuhan}, \textcolor{Italy}{\#spallanzani}, \textcolor{External}{hubei}, \textcolor{External}{\#cinesi}, \textcolor{External}{egitto}, \textcolor{External}{hong kong}, \textcolor{Italy}{civitavecchia}, \textcolor{Italy}{spallanzani} &  External:0.25, Spread:0.20, Italy:0.20, Event:0.10, News:0.05, Solidarity:0.05, Person:0.05, Policy:0.05 \\
Initial      &                           \textcolor{Football}{bresciani}, \textcolor{Football}{\#brescianapoli}, \textcolor{Football}{napoletano}, \textcolor{Football}{coro}, \textcolor{Football}{cantano}, \textcolor{Italy}{cava}, \textcolor{Football}{napoletani}, \textcolor{Italy}{\#codogno}, \textcolor{Spread}{contatto con}, \textcolor{Italy}{codogno} &                                                    \textcolor{Italy}{codogno}, \textcolor{Italy}{castiglione}, \textcolor{Spread}{38enne}, \textcolor{Italy}{\#lodi}, \textcolor{External}{princess}, \textcolor{Italy}{\#codogno}, \textcolor{Italy}{adda}, \textcolor{Spread}{38enne ricoverato}, \textcolor{Spread}{casi in lombardia}, \textcolor{Policy}{quarantena obbligatoria} &                                                                                                           \textcolor{Spread}{38enne}, \textcolor{Italy}{\#codogno}, \textcolor{Italy}{\#lodi}, \textcolor{Italy}{adda}, \textcolor{Spread}{38enne ricoverato}, \textcolor{Policy}{quarantena obbligatoria}, \textcolor{Spread}{casi in lombardia}, \textcolor{Italy}{nel lodigiano}, \textcolor{Italy}{castiglione}, \textcolor{Spread}{contagiati in lombardia} &                                        Italy:0.40, Football:0.30, Spread:0.20, External:0.05, Policy:0.05 \\
Northern     &                                        \textcolor{Spread}{amuchina}, \textcolor{Policy}{scuole chiuse}, \textcolor{Italy}{catania}, \textcolor{Person}{ceriscioli}, \textcolor{Policy}{vertice}, \textcolor{Spread}{\#amuchina}, \textcolor{Italy}{\#palermo}, \textcolor{Italy}{palermo}, \textcolor{Italy}{dal nord}, \textcolor{Person}{emiliano} &                                                   \textcolor{Italy}{\#coronaviruslombardia}, \textcolor{Policy}{scuole chiuse}, \textcolor{Policy}{porte}, \textcolor{Policy}{chiuse}, \textcolor{Spread}{amuchina}, \textcolor{Spread}{panico}, \textcolor{Policy}{\#zonerosse}, \textcolor{Football}{inter}, \textcolor{Spread}{\#coronavirus19italia}, \textcolor{Policy}{rinviata} &                                                                                                                              \textcolor{Italy}{\#coronaviruslombardia}, \textcolor{Policy}{scuole chiuse}, \textcolor{Spread}{amuchina}, \textcolor{Policy}{porte}, \textcolor{Policy}{chiuse}, \textcolor{Policy}{rinviata}, \textcolor{Policy}{\#zonerosse}, \textcolor{Spread}{panico}, \textcolor{Policy}{scuole}, \textcolor{Spread}{\#coronavirus19italia} &                                          Policy:0.35, Spread:0.25, Italy:0.25, Person:0.10, Football:0.05 \\
National     &  \textcolor{Solidarity}{\#restiamoacasa}, \textcolor{Policy}{carabinieri}, \textcolor{Italy}{messina}, \textcolor{Italy}{siracusa}, \textcolor{Solidarity}{\#iostoacasa}, \textcolor{Person}{musumeci}, \textcolor{Italy}{\#sardegna}, \textcolor{Policy}{\#italiazonarossa}, \textcolor{Solidarity}{\#andratuttobene}, \textcolor{Italy}{benevento} &  \textcolor{Solidarity}{\#restiamoacasa}, \textcolor{Solidarity}{\#iorestoacasa}, \textcolor{Solidarity}{\#iostoacasa}, \textcolor{Policy}{autocertificazione}, \textcolor{Solidarity}{\#andratuttobene}, \textcolor{Policy}{\#italiazonarossa}, \textcolor{Event}{pasqua}, \textcolor{Solidarity}{\#stayathome}, \textcolor{Event}{aprile}, \textcolor{Spread}{\#coronavirus19italia} &                                                                                         \textcolor{Solidarity}{\#restiamoacasa}, \textcolor{Solidarity}{\#iostoacasa}, \textcolor{Solidarity}{\#iorestoacasa}, \textcolor{Policy}{\#italiazonarossa}, \textcolor{Solidarity}{\#andratuttobene}, \textcolor{Policy}{autocertificazione}, \textcolor{Event}{pasqua}, \textcolor{Solidarity}{dona}, \textcolor{Solidarity}{\#stayathome}, \textcolor{Policy}{spesa} &                            Solidarity:0.40, Policy:0.20, Italy:0.20, Event:0.10, Person:0.05, Spread:0.05 \\
Prolongation &                                        \textcolor{Italy}{siracusa}, \textcolor{Policy}{\#4maggio}, \textcolor{Policy}{\#congiunti}, \textcolor{Italy}{\#taranto}, \textcolor{Event}{25 aprile}, \textcolor{Event}{\#25aprile}, \textcolor{Italy}{sassari}, \textcolor{Event}{\#pasqua}, \textcolor{Policy}{asporto}, \textcolor{Policy}{per la fase} &                                                                                           \textcolor{Event}{pasqua}, \textcolor{Event}{maggio}, \textcolor{Policy}{\#mes}, \textcolor{Solidarity}{\#stayhome}, \textcolor{Spread}{rsa}, \textcolor{Policy}{riaprire}, \textcolor{Event}{\#25aprile}, \textcolor{Policy}{\#fase2}, \textcolor{Person}{arcuri}, \textcolor{Policy}{fase} &                                                                                                                                               \textcolor{Event}{pasqua}, \textcolor{Solidarity}{\#stayhome}, \textcolor{Event}{\#25aprile}, \textcolor{Policy}{\#4maggio}, \textcolor{Event}{maggio}, \textcolor{Event}{\#pasqua}, \textcolor{Person}{arcuri}, \textcolor{Event}{25 aprile}, \textcolor{Solidarity}{\#stayathome}, \textcolor{Policy}{\#fasedue} &                            Policy:0.40, Event:0.30, Italy:0.15, Solidarity:0.05, Spread:0.05, Person:0.05 \\
Relaxing     &                                       \textcolor{Spread}{covid free}, \textcolor{Italy}{fvg}, \textcolor{Italy}{\#bari}, \textcolor{Italy}{\#basilicata}, \textcolor{Italy}{\#puglia}, \textcolor{Italy}{\#sicilia}, \textcolor{Italy}{caserta}, \textcolor{Spread}{zero contagi}, \textcolor{Spread}{nessun decesso}, \textcolor{Italy}{basilicata} &                                              \textcolor{Spread}{il bollettino di}, \textcolor{Person}{crisanti}, \textcolor{None}{webinar}, \textcolor{Policy}{durante il lockdown}, \textcolor{Person}{zangrillo}, \textcolor{Spread}{seconda ondata}, \textcolor{External}{\#brasile}, \textcolor{Policy}{\#fase3}, \textcolor{Policy}{dopo il covid}, \textcolor{External}{brasile} &                                                                                                                 \textcolor{Italy}{mondragone}, \textcolor{Spread}{covid free}, \textcolor{Person}{zangrillo}, \textcolor{Event}{post covid}, \textcolor{Policy}{dopo il covid}, \textcolor{Policy}{dopo il lockdown}, \textcolor{Spread}{zero decessi}, \textcolor{Policy}{\#fase3}, \textcolor{Spread}{nessun decesso}, \textcolor{Policy}{durante il lockdown} &                                          Italy:0.35, Spread:0.25, Policy:0.15, Person:0.10, External:0.10 \\
Clusters     &                                               \textcolor{Italy}{\#bari}, \textcolor{Italy}{\#taranto}, \textcolor{Italy}{\#abruzzo}, \textcolor{Italy}{\#sardegna}, \textcolor{Italy}{caserta}, \textcolor{Italy}{irpinia}, \textcolor{Italy}{\#basilicata}, \textcolor{Italy}{\#sicilia}, \textcolor{Solidarity}{flash}, \textcolor{Italy}{salerno} &                                                                                            \textcolor{News}{\#ultime}, \textcolor{Football}{\#asroma}, \textcolor{Italy}{liguria}, \textcolor{None}{\#della}, \textcolor{Italy}{\#lazio}, \textcolor{Italy}{romagna}, \textcolor{Italy}{firenze}, \textcolor{Italy}{\#toscana}, \textcolor{Italy}{emilia}, \textcolor{Italy}{piemonte} &  \textcolor{Italy}{italia}, \textcolor{Spread}{emergenza}, \textcolor{Spread}{morti}, \textcolor{Italy}{\#coronavirusitalia}, \textcolor{Policy}{lockdown}, \textcolor{Spread}{virus}, \textcolor{Spread}{casi}, \textcolor{Italy}{lombardia}, \textcolor{Spread}{contagi}, \textcolor{External}{cina}, \textcolor{Italy}{\#covid19italia}, \textcolor{Spread}{positivi}, \textcolor{Italy}{roma}, \textcolor{News}{aggiornamento}, \textcolor{External}{cinesi} &                                                                                                           \\
Ratios       &                                                                                                                                                                                                                                             Italy:0.39, Spread:0.17, Policy:0.14, Football:0.10, Event:0.07, Solidarity:0.07, Person:0.05, News:0.02 &                                                                                                                                                                                                                                                                           Policy:0.27, Spread:0.17, External:0.14, Italy:0.14, Event:0.10, Solidarity:0.10, Person:0.07, Football:0.02 &                                                                                                                                                                                                                                                                                                                                                                                                                                                                  &                                                                                                           \\

\bottomrule
\end{tabularx}
\end{center}
\caption{The original table we obtained through our analysis with the Italian tokens}
\label{table:fscorecomparisonitaly}
\end{table}

\end{document}